\documentclass{article}

\usepackage{arxiv}
\usepackage{wrapfig}
\usepackage{graphicx}
\usepackage[utf8]{inputenc} 
\usepackage[T1]{fontenc}    
\usepackage{hyperref}       
\usepackage{url}            
\usepackage{booktabs}       
\usepackage{amsfonts}       
\usepackage{nicefrac}       
\usepackage{microtype}      
\usepackage{lipsum}
\usepackage[ruled,vlined]{algorithm2e}

\title{Listwise Learning to Rank with Deep Q-Networks}

\author{
  Abhishek Sharma\\
  University of California, Berkeley
}

\begin{document}
\maketitle

\begin{abstract}
\textbf{Learning to Rank is the problem involved with ranking a sequence of documents based on their relevance to a given query. Deep Q-Learning has been shown to be a useful method for training an agent in sequential decision making \cite{dqn_paper}. In this paper, we show that DeepQRank, our deep q-learning to rank agent, demonstrates performance that can be considered state-of-the-art. Though less computationally efficient than a supervised learning approach such as linear regression, our agent has fewer limitations in terms of which format of data it can use for training and evaluation. We run our algorithm against Microsoft's LETOR listwise dataset \cite{letor} and achieve an NDCG@1 (ranking accuracy in the range $[0,1]$) of 0.5075, narrowly beating out the leading supervised learning model, SVMRank (0.4958). 
}

\end{abstract}
\section{Introduction}
In the Learning to Rank (LTR) problem, we are given a set of queries. Each query is usually accompanied by multiple (hundreds of) "documents", which are items with varying degrees of relevance to the query. These document-query pairs are common in search engine settings such as Google Search, as well as recommendation engines. The goal of LTR is to return a list of these pairs so that the documents are n intelligently ranked by relevance. Most models approximate a relevance function $f(X) \rightarrow Y$, where $Y$ is the "relevance score" for document $X$. This is known as $pointwise$ learning to rank. These models require the dataset to include a predetermined score that accompanies each document-query pair. In $listwise$ learning to rank (the focus of this project), the document-query pairs have no target value to predict, but rather a correct order in which the documents are pre-ranked. The model's job is to reconstruct these rankings for future document-query pairs.

\begin{figure}[h]
        \centering
        \includegraphics[scale=0.25]{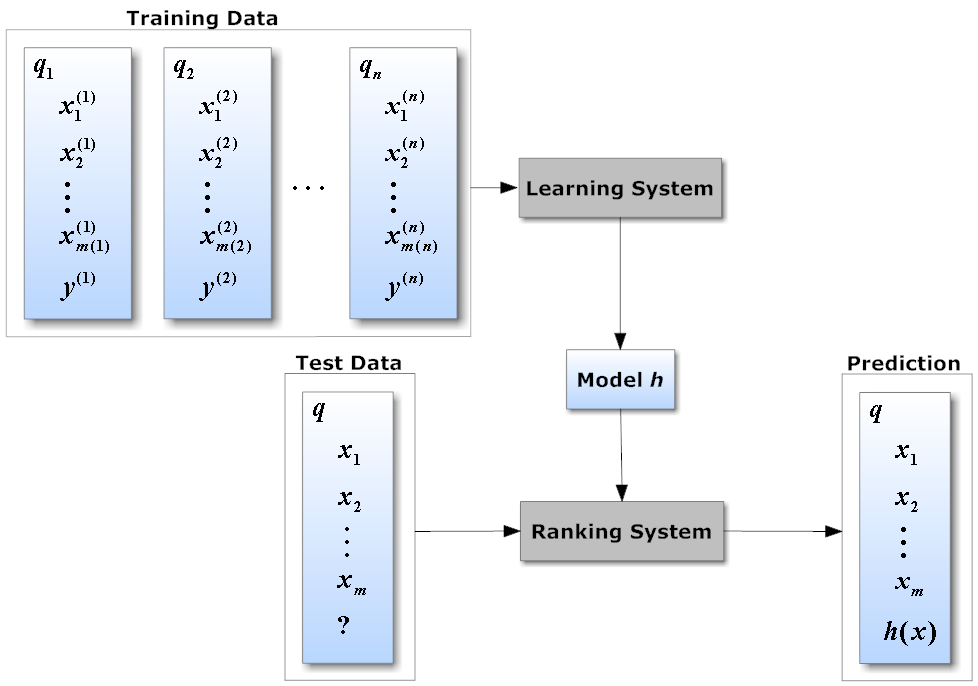}
        \caption{Example workflow for RankNet, a neural network for LTR used in Microsoft's Bing search engine}
        \label{fig:my_label}
\end{figure}

In this work, we propose DeepQRank, a deep q-learning approach to this problem. Related work has shown the effectiveness of representing LTR as a Markov Decision Process (MDP). We build upon this research by applying a new learning algorithm to the application of ranking. In order to apply deep q-learning, we must first represent Learning to Rank as an MDP. Essentially, the agent begins with a randomly ordered list of documents. It selects documents (the action) from this list one by one based on which has the maximum estimated reward. Equipped with this formulation, our agent is now ready to rank documents.

In relevant research documents \cite{source1}, we see that when formulated as an MDP problem, teams have out-performed state of the art baselines for this problem using policy gradients. Deep Q-Learning can benefit from this MDP representation. This is because the problem of generating a ranking is sequential in nature. Most current methods try to optimize a certain metric over the output list. Known as pointwise methods, these programs generate a list that is sorted based on a predicted value for each element. They require the metrics that are being optimized to be continuous, whereas using a ranking evaluation measure for the reward function allows us to optimize the reward without worrying about it being differentiable. 

With our MDP representation, we propose DeepQRank, a deep q-learning agent which learns to rank. Before we apply deep q-learning, we make a few modifications to the classic algorithm. Our algorithm randomly samples a mini-batch from a given Buffer. Next, it computes a target value for the batch based on the actual reward of the action in the batch and the next state. Finally, it updates the target Q-network using this target value. Instead of updating after an arbitrary number of iterations, we introduce a running average (similar to Polyak averaging) to update the weights after every iteration. After building and training the agent, we test it against a special benchmarking dataset. 

\section{Related Work}
This project was inspired by the work of a team at Institute of Computing Technology, Chinese Academy of Sciences \cite{source1}, as well as a team from Alibaba \cite{source3}. Their work on framing learning to rank as a reinforcement learning problem was useful in setting up this experiment. In the first paper \cite{source1}, they express the ranking task as an MDP. Each state consists of the set of unranked, remaining documents, while actions involve selecting a single document to add to a ranked list. The second group takes this representation one step further by building a recommendation engine with MDP Ranking. Another related project from Boston University \cite{source4} deals with the application of a deep neural network for ranking items in a list. This neural network learns a metric function to generate a sort-able value for ranking documents in $O(n\log n)$ time.

While both MDP representations and deep learning approaches performed well in their approaches to LTR, each has its limitations. The MDP approach isn't able to learn a complex function to represent a document's rank since it isn't paired with a neural network. Meanwhile, the deep learning approach requires each document to have a "relevance" label in a discrete range. For example, some datasets assign a relevance strength in the range $[0,4]$ to each document-query pair. This makes the dataset suitable for supervised learning approaches such as the neural network called "FastAP" \cite{source4}. We hope to build a method that serves as advantageous compared to solely using reinforcement learning or deep learning. We believe this because a target neural network network can learn a more complicated relevance function than a method such as a Support Vector Machine.

\section{Learning to Rank as a Markov Decision Process}
In order to apply Deep Q-Learning, we need to express Learning to Rank in the form of an MDP.
Here are the definitions for the state, action, transitions, reward, and policy: \cite{source1}
\begin{enumerate}
    \item \textbf{State:} A state consists of 3 elements: timestep $t$, a set of ranked documents $Y$ (initially empty), and a set of unordered documents $X$ from which we must create our final ranked list. A terminal state is achieved when $X$ is empty. An initial state has empty $Y$ and timestep $t = 0$. 
    \item \textbf{Action:} Our agent can take an action $a_t$ at timestep $t$. The action simply consists of a document $d$ from our unranked set $X$. When the agent performs an action, $d$ is removed from $X$ and added to $Y$. Also, the timestep $t$ is incremented by one. In many traditional MDP formulations, the action set is state-independent. For example, in Pac-man, the action set is consistently [up, down, left, right]. In LTR, this is not the case, as the action set is dependent on which documents are in the state's remaining list.
    \item \textbf{Transition:} Our transition $T$ maps a state, $s$, and action, $a$, to a new state $s'$. These three elements $(s,a,s')$ compose a transition. 
    \item \textbf{Reward:} Every state-action pair has a corresponding reward. We developed the following formula based off Discounted Cumulative Gain:
    $$r_{s, a} = \frac{\textbf{rank}(doc_a)}{\log_2(t_{s}+1)}$$
    In this equation, $\textbf{rank}(doc_a)$ gives the rank value (between 1 and $|query_s|$) of a document selected in action $a$ (higher ranking documents have stronger relevance to the query), and $t_s$ is the timestep from state $s$. We must add a $+1$ in the denominator's log statement to prevent a division by zero (initial states have a timestep of zero). Here, we penalize the selection of high-ranking documents late in the ranking process. In order to maximize reward, the agent will have to select the highest ranking documents as early as possible. This reward function is one area that warrants further exploration. Since there are many other metrics for measuring ranking accuracy, such as Kendall's Tau and the Spearman Rank Correlation, it may be a point of future interest to experiment with varying reward functions to see how they affect the agent's accuracy.
    \item \textbf{Policy:} Our policy $P: S \rightarrow A$ maps any given state to an action $A$. The agent runs its neural network on each document in the state's \textit{remaining} list. It then returns the document with the highest reward, as estimated by the network. 
\end{enumerate}
Ideas for this representation are heavily influenced by the work of Wei et al \cite{source1}. Some modifications were made to include the ranked list in state representations, as well as a larger scale of rank values in the reward function. Additionally, from an engineering standpoint, the state representation was modified to include query id, which is useful when training the agent on multiple queries at once.

Under the MDP formulation, here's how a trained RL agent would rank a set of documents $X$, based on their relevance to a query $q$.

\begin{algorithm}[H]
\SetAlgoLined
\KwResult{A ranked list $Y$}
 \textbf{Input:} Trained Agent $\mathcal{A}$, Unranked list $X$\;
 set $Y = []$\;
 set timestamp $t = 0$\;
 set current state $S = State(t, Y, X)$\;
 \While{length of X > 0}{
 Run the forward pass of $\mathcal{A}'s$ model with the current state and every document in X\;
 Remove the document $X_i$ with the highest output from X and add it to $Y$\;
 t++\;
 Set current state = State(t, Y, X)\;
 }
 return $Y$\;
 \caption{GetRanking Function for DeepQRank}
\end{algorithm}

\section{Modifications to "Classic" Deep Q-Learning}

\subsection{Weighted Average}
The main change to the deep q-learning algorithm is the use of polyak averaging. While the classic deep q-learning algorithm updates the network parameters by copying from the current iteration every $N$ steps, our algorithm updates its parameters slightly every step with the following formula:
$$\phi' \leftarrow \tau \phi' + (1-\tau) \phi$$
, where $\phi'$ is the target network parameters, $\tau$ is a chosen value ($0.999$ works well in practice), and $\phi$ is the network parameters at the end of the current step.

\subsection{DeepQRank}

The previously stated modifications to the agent and underlying model result in the following modified learning algorithm for our Deep Q-Network:

\begin{algorithm}[H]
\SetAlgoLined
\KwResult{A trained Q-network with parameters $\phi'$}
 \textbf{Input:} Number of steps $\mathcal{S}$, Buffer $\mathcal{B}$\;
 set $E \leftarrow EpisodeSample()$ (from algorithm in $5.2$) \;
 initialize $\phi'$
 int i = 0\;
 \While{$i < \mathcal{S}$}{
  sample minibatch $mb_i$ from $\mathcal{B}$ uniformly\;
  compute $y_i \leftarrow r_i + \gamma max_{a'_{i}}Q_{\phi'}(s'_i, a'_i)$ \;
  $\phi \leftarrow \phi - \alpha \sum_i \frac{dQ_{\phi}}{d\phi}(s_i, a_i)(Q_{\phi}(s_i, a_i)-y_i)$\;
  $\phi' \leftarrow \tau \phi' + (1-\tau) \phi$\;
  $i \leftarrow i + 1$\;
  }
  return $\phi'$\;
\caption{Deep Q-Learning to Rank Algorithm}
\end{algorithm}

We implemented this algorithm in Python (see appendix) and observed promising results with the LETOR dataset.

Here are the specs for the DeepQRank agent:
\begin{itemize}
    \item Target Network Architecture: Our neural network is fully connected with the following layer sizes: \\ 
    47 (input layer), 32 (hidden layer), 16 (hidden layer), 1 (output layer).\\
    \begin{figure}[h]
        \centering
        \includegraphics[scale=0.3]{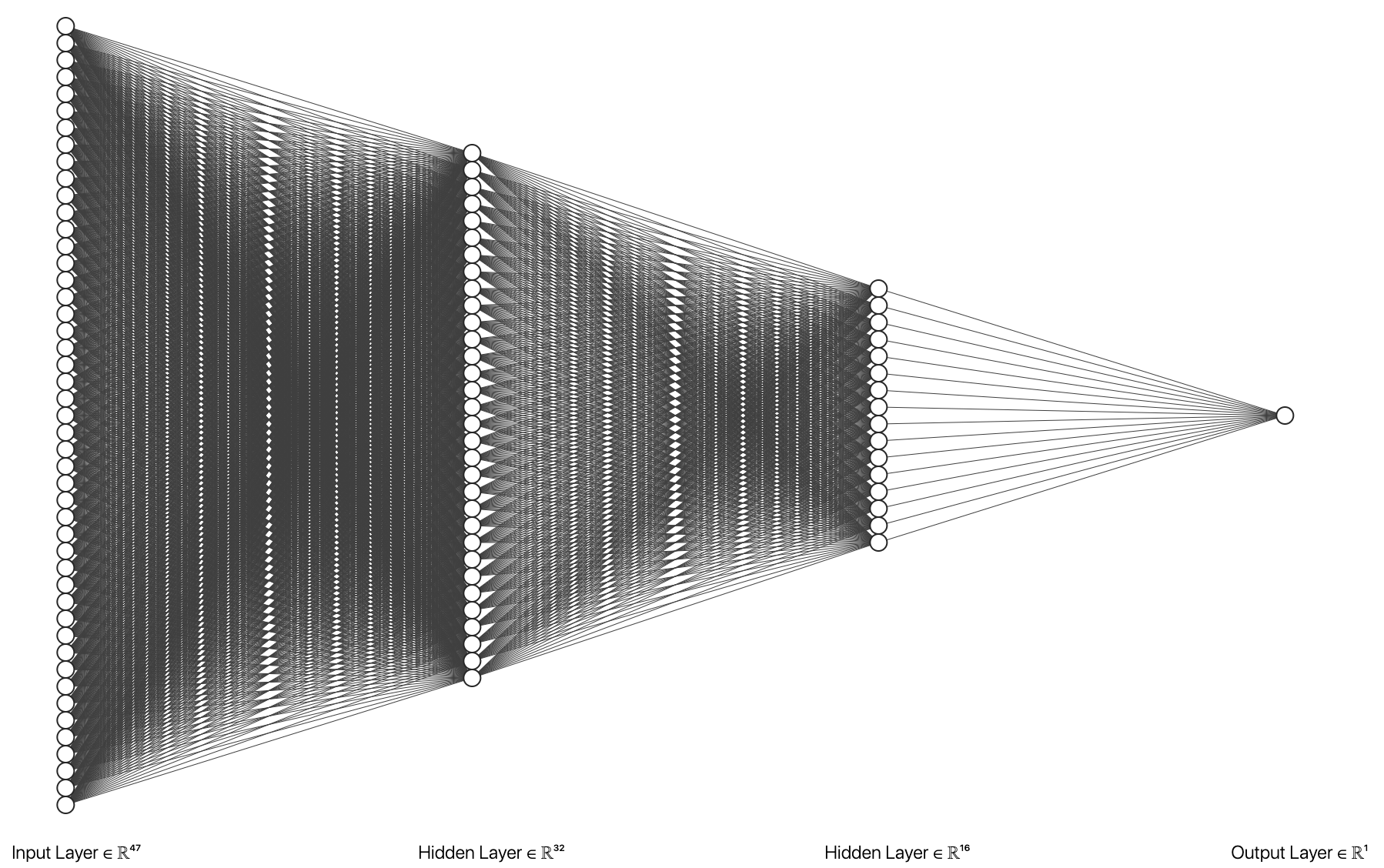}
        \caption{Feedforward Neural Network Architecture for our Agent}
        \label{fig:my_label}
    \end{figure}
    \item Learning Rate $\alpha$: $3*10^{-4}$
    \item Discount factor $\gamma$: 0.99
    \item Polyak averaging factor $\tau$: 0.999

\end{itemize}

\subsection{Comparative Analysis of DeepQRank}
\subsubsection{Advantage: No "Relevance" Needed}
With most supervised learning methods for LTR, the dataset includes a "relevance" label which the model tries to predict. For example, a subset of the LETOR dataset includes labels in the range $[0,4]$. A classifier can be trained to classify a document into one of these 5 classes using the document's features. Unfortunately, classifiers like these are not compatible with datasets that present documents in order of relevance (without specific relevance labels in the $[0,4]$ scale). This makes DeepQRank more suitable for listwise learning to rank, in which the agent learns from a ranked list rather than a set of target relevance labels. 
\subsubsection{Disadvantage: Runtime analysis}
While the neural network approach in "Deep Metric Learning to Rank" can rank a list in $O(n\log n)$ time, this algorithm would take at least $O(n^2)$ time. This is because the neural network can compute the "metric" measure for every document in a single forward pass, and then simply sort the documents by this metric with an algorithm such as MergeSort. Meanwhile, DeepQRank computes a forward pass on the entire batch of remaining documents every time it picks the next document to add to its running ranked list. Both of these runtime analyses assume roughly equivalent times for their neural network forward passes, so that element is simplified to $O(1)$ time.

\section{Experiment and Results}

\subsection{Data}
We used the LETOR listwise ranking dataset for this project. In this dataset, each row represents a query-document pair. The headers for the dataset consist of the following: query id, rank label, $feature_1, feature_2, ..., feature_{46}$. Query ID identifies which query was requested for a given pair. Rank label corresponds to a document's relevance for the particular query in its pair. A higher rank label signifies stronger relevance. Maximum values for rank label depend on the size of the query. Therefore, the challenge here is in predictive relative relevance values for ranking, rather than just turning this into a regression problem for predicting relevance. 

One advantage that this dataset serves over traditional learning to rank datasets is that it is "fully ranked". What this means is that for a given query, the dataset provides an exact ranking of all of the corresponding documents.  Most datasets for supervised learning to rank generalize the ranking with provided "target values" that come in small ranges. This modification caters to approaches such as linear regression.

For example, a larger dataset in the LETOR collection includes query-document pairs, but substitutes a "relevance label" for a rank. This relevance label is in the range $[0,4]$. For a query with $1,000$ datasets, this means that there are multiple "correct" rankings, as two documents $a,b$ with a relevance label of $4$ can be ordered either way and both be considered correct. While these datasets have been produced to accomodate supervised learning approaches to ranking such as multivariate regression and decision trees, they don't paint the full picture of a definitive ranking such as the LETOR listwise dataset.

\subsection{Experiment Setup}
Before running and evaluating our algorithm, we needed to setup the buffer. Buffer collection requires an algorithm of its own because of the nature of the dataset we are working with. 

\begin{algorithm}[H]
\SetAlgoLined
\KwResult{$E$, an "episode" consisting of $M$ tuples $(s_t, a_t, s_{t+1}, r_{t+1})$ for a length $M$ query}
 set $E\leftarrow$ []\;
 dataset $\mathcal{D}$ (contains rows with document-query pair information)\;
 string $Q\leftarrow$ random query id from $\mathcal{D}$\;
 set $X\leftarrow$ all rows of $\mathcal{D}$ with query id = $Q$\;
 set $Y\leftarrow$ []\;
 int $t\leftarrow 0$\;
 set initial state $S \leftarrow$ $X, Y, t$\;
 \While{$X$ is not empty}{
  save oldState $\leftarrow$ S\;
  row $R \leftarrow$ pop a random row from $X$\;
  action $a_t \leftarrow$ document id of row $R$\;
  append $R$ to $Y$\;
  $t \leftarrow t+1$\;
  reward $r_t \leftarrow$ reward $r_{S, a_t}$ (defined in section 3)\;
  update $S \leftarrow$ $X, Y, t$\;
  append (oldState, $a_t, S, r_t$) to $E$

 }
 \caption{LETOR Episode Sample for a Single Query $Q$}
\end{algorithm}

\subsection{Visualizations}

We tracked 2 performance-related variables over time: the loss of our agent's target network and the actual validation accuracy, measured using the NDCG@1 metric on an isolated validation set.

\begin{enumerate}
    \item \textit{Model Loss}: Mean Squared Error loss for the agent's neural network. This is the first measure that we observed over time to ensure improvement for our model.
    
    \begin{figure}[h]
        \centering
        \includegraphics[scale=0.7]{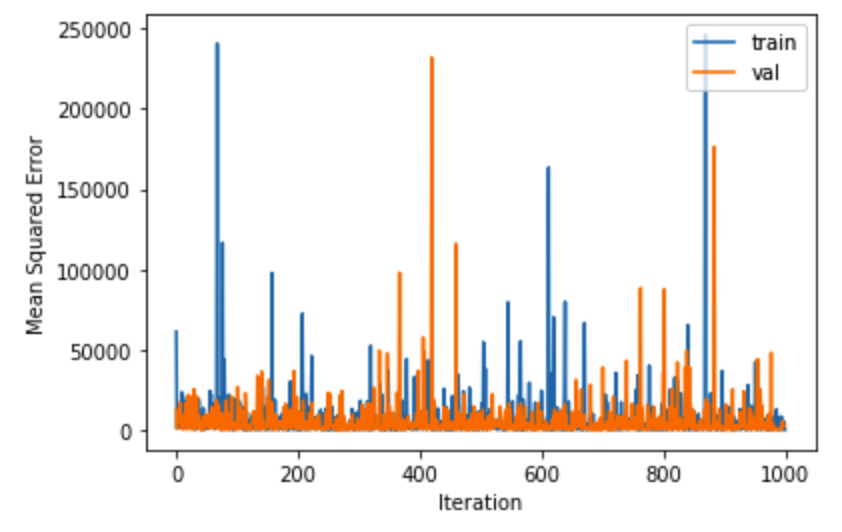}
        \caption{Raw mean squared error over time for our target network.}
        \label{fig:my_label}
    \end{figure}
    
    At first glance, this plot paints the picture of a stagnated model that isn't learning. After manipulating the data, we're able to better visualize the model's behavior by observing a moving average of $\log_{10}$ (applied to smooth out the large range of values) loss instead of plain MSE loss.
    
    \begin{figure}[h]
        \centering
        \includegraphics[scale=0.45]{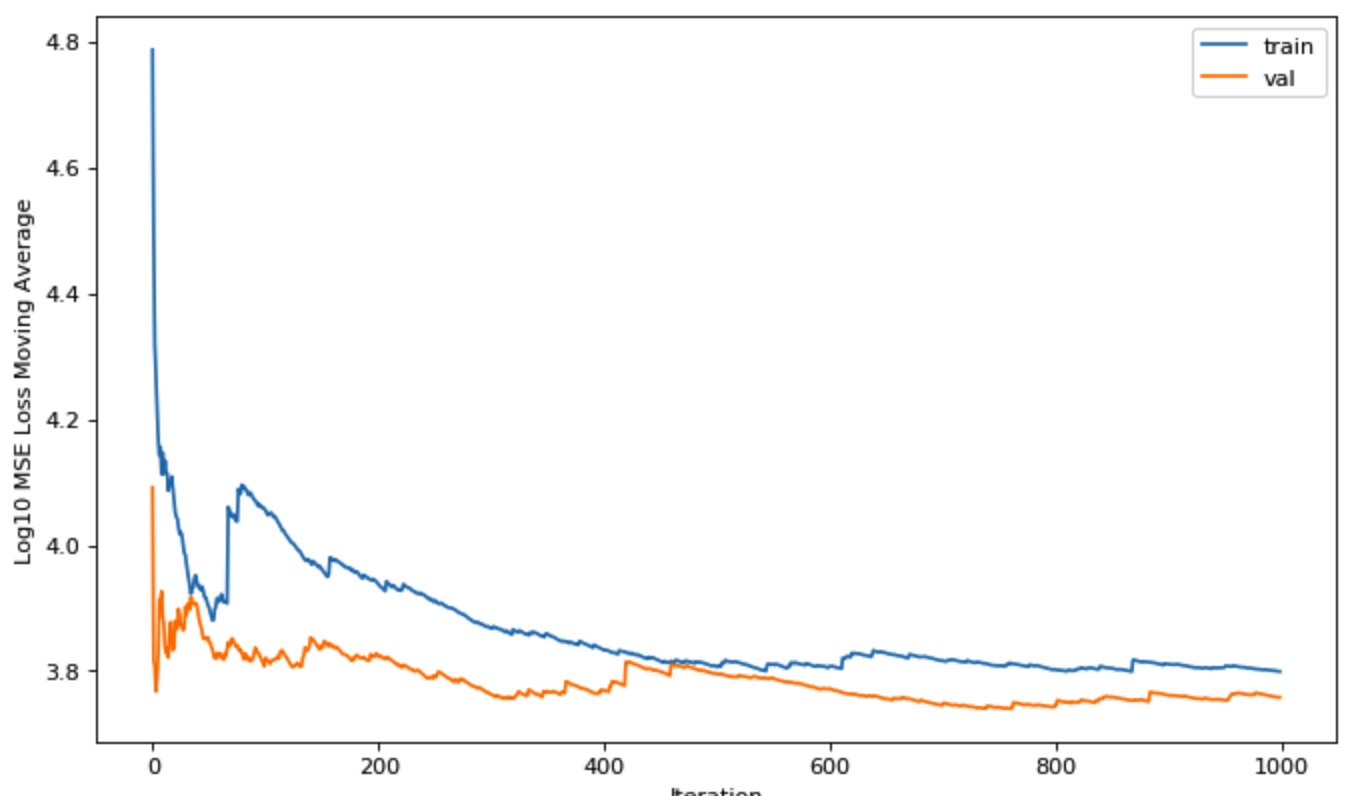}
        \caption{A moving average of $\log_{10}$ loss over time}
        \label{fig:my_label}
    \end{figure}
    
    After initially jumping upward, the network begins to achieve a stable learning pattern that slowly minimizes loss. We believe the initial jump is due to an outlier which skews the moving average within the first 10 iterations of training. The validation loss and training loss both improve at similar rates. This is evidence that the model is not overfitting on the training data. One possible explanation for training loss being higher than validation loss is the introduction of an outlier in the training set which spikes the training loss moving average upward. This in turn affects every moving average after. The importance attribute to notice here is that both steadily decrease over time, providing evidence that the model is learning from the samples in the training buffer.
    
    \item $NDCG @ k$: Normalized Discounted Cumulative Gain at position $k$. This was the main metric used to evaluate the agent's performance in ranking. This formula was referenced in our definition of the reward. With the following definitions for $DCG_K$ and $IDCG_k$:
    $$DCG_k = \sum_{i=1}^{p} \frac{2^{rel_i}-1}{\log_2(i+1)}, IDCG_k = \sum_{i=1}^{|REL_k|} \frac{2^{rel_i}-1}{\log_2(i+1)}$$
    $NDCG @ k$ is calculated as follows:
    $$nDCG_k = \frac{DCG_k}{IDCG_k}$$
    
    Rankings with perfect accuracy would score an NDCG@k of $1$, while a really weak ranking scores $0$. 
     Most well-trained model achieve an NDCG@1 value between $0.3$ and $0.5$.
     For benchmarking purposes, we specifically computed $NDCG @ 1$.
    
    Here's how average NDCG@1 on our validation set improved over time:
    
    \begin{figure}[h]
        \centering
        \includegraphics[scale=0.7]{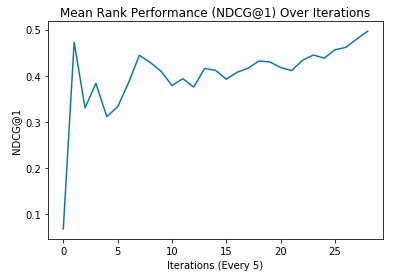}
        \caption{Our primary ranking accuracy metric, NDCG@1, over time.}
        \label{fig:my_label}
    \end{figure}
    
    As the graph shows, the first few iterations sparked a drastic improvement. After about 50 iterations (x=10 in the plot), we see a stabilized improvement over time. At the end of 150 iterations, the agent scored an NDCG@k of $0.5075$, beating out many state-of-the-art techniques.
\end{enumerate}

\subsection{Relative Ranking Performance}

At the end of our training period, we ran a final evaluation on the DeepQRank agent on our test set. This returned a mean NDCG@1 value of $0.5075$.

\begin{center}
\begin{tabular}{|l|l|}
\hline
\multicolumn{2}{|c|}{NDCG@1 for various ranking models} \\
\hline
RankSVM      & 0.4958     \\
ListNet      & 0.4002     \\ 
AdaRank-MAP  & 0.3821     \\
AdaRank-NDCG & 0.3876     \\
SVMMAP       & 0.3853     \\ 
RankNet       & 0.4790     \\ 
MDPRank      & 0.4061     \\ 
\textbf{DeepQRank}      & \textbf{0.5075}     \\
\hline
\end{tabular}
\end{center}

While all other models score below $0.5$ with the NDCG @ 1 measure, DeepQRank is the only one which scores above $0.5$. Though this is a preliminary experiment, these results signify that DeepQRank is a promising method which should be investigated further to verify its effectiveness.

We speculate that RankSVM is limited in its ability to rank because it doesn't have the ability to approximate the relevance function as well as a deep neural network. In its linear form, it's even more limited since it can't approximate a nonlinear relevance function.

\section{Conclusion}

In summation, we modified deep q-learning by customizing the reward function and neural network architecture, and introduced a polyak averaging-like method in the training phase. Our experiment saw successful results when measured with the Normalized Discounted Cumulative Gain metric. Based on the outcome of this investigation, we are enthusiastic about further researching and improving this learning to rank method.

For future work, there are quite a few potential avenues for replicating and strengthening the results of this investigation. First, it may be useful to incorporate more datasets and evaluate their performance against official benchmarks that have been recorded for competing ranking methods. For instance, the Yahoo webscope dataset, though only providing relevance labels in the range $[0,4]$, contains 700 features per document-query pair. This may improve the performance of our deep q-network. We may want to change the architecture of the target network itself. It's possible that adding more neurons / hidden layers or introducing modern features such as dropout improve the neural network's loss. 
Lastly, there are many more deep reinforcement learning methods that have yet to be applied to the MDP representation of this problem. To our knowledge, policy gradient and deep q-learning are the only such methods to be tested so far.

It's likely that the formulas in this paper can be tuned to improve performance. For example the reward function, based on Discounted Cumulative Gain, may be modified to yield better results. The current $\log_2$ denominator is a variable that could modified to change the distribution of the observed reward.

\bibliographystyle{unsrt}  


\end{document}